\documentclass{article}

\usepackage{arxiv}
\usepackage{eurosym}
\usepackage[utf8]{inputenc} 
\usepackage[T1]{fontenc}    
\usepackage{hyperref}       
\usepackage{url}            
\usepackage{booktabs}       
\usepackage{amsfonts}       
\usepackage{nicefrac}       
\usepackage{microtype}      
\usepackage{lipsum}
\usepackage{graphicx}
\usepackage{xcolor}
\usepackage{amssymb}
\usepackage{subcaption}
\usepackage{amsmath}
\usepackage[autostyle]{csquotes}
\usepackage{algorithm}
\usepackage{algorithmic}
\usepackage{textcomp}
\usepackage{trimclip}
\usepackage{multirow}
\usepackage{adjustbox}
\usepackage{hhline,colortbl}
\usepackage{atbegshi}
\usepackage{verbatim}
\usepackage{array}
\usepackage{lineno,hyperref}
\usepackage{breakurl}

\usepackage{subcaption}
\usepackage{fouriernc} 
\usepackage{soul}
\usepackage{diagbox}
\usepackage[inline]{enumitem}
\usepackage{longtable}
\usepackage{ltxtable}
\usepackage{tabularx}
\usepackage{dblfloatfix}
\usepackage{lscape}
\usepackage{tikz}
\usepackage{pgfplots}

\title{The emergence of Explainability of Intelligent Systems: Delivering Explainable and Personalised Recommendations for Energy Efficiency}

\author{Christos Sardianos \thanks{This paper has been accepted in International Journal of Intelligent Systems}, Iraklis Varlamis, Christos Chronis, George Dimitrakopoulos \\
Department of Informatics and Telematics\\
Harokopio University of Athens\\
Athens, Greece\\
\texttt{sardianos@hua.gr;varlamis@hua.gr;it21797@hua.gr;gdimitra@hua.gr} \\
  \And
Abdullah Alsalemi, Yassine Himeur, Faycal Bensaali\\
  Department of Electrical Engineering\\
  Qatar University\\
  Doha, Qatar \\
  \texttt{yassine.himeur@qu.edu.qa;a.alsalemi@qu.edu.qa;f.bensaali@qu.edu.qa} \\
   \And
 Abbes Amira \\
  Institute of Artificial Intelligence\\
  De Montfort University\\
  Leicester, United Kingdom \\
  \texttt{abbes.amira@dmu.ac.uk} \\
}

\begin{document}
\maketitle

\begin{abstract}
The recent advances in artificial intelligence namely in machine learning and deep learning, have boosted the performance of intelligent systems in several ways. This gave rise to human expectations, but also created the need for a deeper understanding of how intelligent systems think and decide. The concept of explainability appeared, in the extent of explaining the internal system mechanics in human terms. Recommendation systems are intelligent systems that support human decision making, and as such, they have to be explainable in order to increase user trust and improve the acceptance of recommendations. In this work, we focus on a context-aware recommendation system for energy efficiency and develop a mechanism for explainable and persuasive recommendations, which are personalized to user preferences and habits. The persuasive facts either emphasize on the economical saving prospects (Econ) or on a positive ecological impact (Eco) and explanations provide the reason for recommending an energy saving action. Based on a study conducted using a Telegram bot, different scenarios have been validated with actual data and human feedback. Current results show a total increase of 19\% on the recommendation acceptance ratio when both economical and ecological persuasive facts are employed. This revolutionary approach on recommendation systems, demonstrates how intelligent recommendations can effectively encourage energy saving behavior.
\end{abstract}


\section{Introduction}\label{sec_intro}

Currently, it is more evident than ever before that we consume much more energy than we actually need. The extensive use of electrical devices in most of our daily activities, the absence of eco- or energy-friendly design in older devices, and the lack of awareness or care from the consumers' side for reducing their energy footprint leads to an over-consumption of energy. In order to promote energy efficiency awareness and care at the product level and to trigger a global harmonisation, the EU has established a regulatory framework \footnote{https://www.europarl.europa.eu/factsheets/en/sheet/69/efficjenza-energetika}, which among others defines a set of Minimum Energy Performance Standards (MEPS) and an energy-related labelling scheme for electrical devices. In the hypothetical scenario where all countries agree to the MEPS and apply them by 2020, the gross annual energy savings are expected to reach 13\% by 2030, and will reach 34\% if the highest energy efficiency levels are agreed and put into practice \cite{molenbroek2015savings}. These findings show the importance for investing on more efficient electrical devices, but also point a need for adopting a more energy efficient behavior. 

In order to promote better behaviors, and discourage bad ones, governments use several policy interventions such as energy efficiency labeling for devices, taxation of high energy consumption and financial incentives for consumption reduction, which have short or medium effect to consumers' behavior.
They also provide feedback, energy savings tips, and peer device comparisons in order to persuade consumers on the benefits from a behavioral change \cite{cattaneo2019internal}. The aforementioned interventions have a bigger impact when they are addressed to domestic users. However, their impact is smaller in public buildings, such as schools or offices, where people tend to care less for proper energy usage, since they do not directly pay for the consumed energy. The main reason behind high energy consumption in public buildings is the unnecessary usage of devices (e.g. heating or cooling devices, lights, etc.), especially when the public spaces are not occupied \cite{rafsanjani2015review}. 
In this case, it is important either to automate energy efficiency, by embedding intelligence to the devices (actuators) and the environment (sensors), or to gradually change people's habits and promoting more energy efficient behaviors, through warnings, notifications and up-to-time recommendations \cite{kluckner2013exploring, petkov2012personalised, graml2011improving}. 
Despite the many works that used feedback for persuading home users to improve their energy consumption behavior, it was in \cite{alsalemi2019ieeesystems} that action recommendations have been addressed in real-time to home users, in association to their actions and daily habits.

According to the Habit Loop theory \cite{em3_gardner2015review}, the main neurological loop that governs any change in habit comprises: a cue, a routine, and a reward. In order to replace inefficient energy habits with efficient ones it is important to identify the most promising routines for energy saving, locate their cues and recommend energy saving actions. Recommender systems can be very supportive in this change loop, since they can link the action with the reward, and strengthen the new routine.

Generally, recommendation tasks can be classified as addressing the five \textit{W} components: when, where, who, what, why \cite{zhang2020explainable}. The five W’s generally correspond to time-aware recommendations (\textit{when}), location-aware recommendations (\textit{where}), their social aspect (\textit{who}), application-specific recommendations (\textit{what}), and their explainable component (\textit{why}), respectively. In this work, we focus on the \textit{why} aspect with the help of explainable recommender systems.

Following the trend of explainable AI (Artificial Intelligence), the explainable recommender systems aim to provide users with useful recommendations, followed by explanations about them \cite{zhang2020explainable}. Explanations may refer to the reasons behind the recommendation or to the benefits from choosing the recommended option. They can improve the persuasiveness of the system, the user understanding and satisfaction and provide an immediate reward to the user. 

Recent explanatory work focused on two dimensions categorized as: 1) the form of explanations produced (e.g. textual, visual, etc.); and 2) the model or algorithm used to produce the said explanation, which can be loosely categorized as matrix factorisation, topic modeling, graph-driven, deep learning , knowledge-graph, interaction laws, and post-hoc models, etc. \cite{zhang2020explainable}

Explainable recommendations can be classified by the type of explanation used:

\begin{enumerate}
\item User-based and Item-based Explanations
\item Content-based Explanation
\item Textual Explanations
\item Visual Explanations
\item Social Explanations
\item Hybrid Explanations
\end{enumerate}

In this work, we focus on creating hybrid explanations that combine the power of contextual and textual explanations.

This work builds on our previous work on micro-moment based recommendations, and powers them with information on the reasons that triggered each recommendation and the expected benefit from its acceptance. The main contributions of this work comprise:  
\begin{itemize}
\item a recommendation system, which generates personalised recommendations aligned with the user goals;
\item the ability of the system to learn from the user response to a recommended action and adapt the recommendations that follow;
\item the explainability of recommendations both in terms of reasoning the selection of a proposed action and of providing the user with persuasive facts about energy savings from the action.
\end{itemize}

Section \ref{sec_related_work} that follows provides a summary of the related works in the field, whereas Section \ref{sec_methodology} presents the proposed methodology for explainable and persuasive recommendations, which bring human in the loop of an energy efficiency system. The section also presents the core system architecture with emphasis on the explainable recommendation extensions. Section \ref{sec_evaluation} performs a comparative evaluation of the various recommendation strategies in order to demonstrate the improvement in performance and the facts that mostly affect user choices. Finally, Section \ref{sec_conclusions} summarizes the main findings of this work and indicates the next steps.

\section{Related Work}\label{sec_related_work}

In order to better understand and explain the suggestions of recommendation systems for energy efficiency it is first necessary to perform a comprehensive survey of recommendation algorithms. The survey that follows presents the main recommendation techniques and their applications in the domain of energy efficiency, in an attempt to cover the entire depth and breadth of state-of-the-art approaches so far. The surveyed approaches are summarized in Table \ref{RelWorks}.
Then focuses on the explanations and facts that recommendation systems can use in order to improve user acceptance.

\subsection{Recommendation techniques for energy efficiency}

\subsubsection{Case-based}
Case-based recommender systems are rule based systems, which can work for one or more users, by considering each user individually. The individual consumption habits and preferences are evaluated against a set of rules and predefined decisions, which trigger - when met - the corresponding energy saving actions.
Authors in \cite{Bravo2019} implement a multi-agent system, which enables to (i) collect power usage patterns from electrical appliances in domestic buildings; (ii) procure electricity price data from Internet; (iii) trigger appropriate recommendations for end-users using consumption footprints electricity prices. To this end, the developed recommendation system furnishes information about the hours to use domestic devices, offering an economic benefit to end-users. This can be regarded as a strategy to distribute/optimize the use of power in households and avoid peak electricity demand. 
A case-based reasoning recommendation system is introduced in \cite{Pinto2018}. The system knowledge (cases) are historic related examples which map a usage behavior to an energy saving plan. The system recommends an energy-saving plan to a user at each specific moment of the day, by considering his/her consumption behavior and similar cases from the knowledge base. A k-Nearest Neighbor (KNN) technique is used to retrieve the most similar examples at each moment and an SVM-based weighting scheme is employed for optimizing weighting factors of each example. At the last stage, an expert system, which contains an ensemble of ad-hoc rules that ensure the applicability of this strategy to the case is used. The authors develop the aforementioned  case-based reasoning scheme using a committed software agent, which allows the integration of the recommender system in a multi-agent framework with more energy saving capabilities.

\subsubsection{Collaborative filtering}
Collaborative filtering techniques assume a set of users that choose from a closed set of items (or actions) and explicitly or implicitly state their preferences (or ratings) for them. The items recommended to the user are the most preferable to him/her (or those with the highest predicted rating) \cite{morawski2017fuzzy} or to the group of users he/she belongs to\cite{castro2018group} .
Energy saving systems that employ collaborative filtering  deploy different interacting intelligent agents which dynamically capture user preferences. They then promote energy-efficiency to end-users through tailored recommendations that better match their preferences. 

More specifically, the authors of \cite{Zhang8412100} focus on the analysis of house appliances data and try to predict the rating levels of various consumption plans, which correspond to the user preferences for each plan. Then, they use the prediction model to help new users to select pertinent plans and appropriate tariffs.
The energy-saving recommendations in \cite{ZHENG2020117775} are generated using a dual-step framework, which involves a feature formulation and a recommendation generation phase. In the former phase, user preferences are captured in a matrix, in which the rows correspond to energy-saving devices and the columns correspond to users. The matrix values depict the appliance usage information for each user and each device. As a result, the matrix models end-users' consumption behavior and users are represented as vectors in the feature space of devices. The collaborative filtering algorithm in the second phase, performs a kNN clustering of users, which is the basis for sending the same tailored recommendations to all users of the same cluster.  
The ReViCEE recommendation system \cite{KAR2019135} provides personalized recommendations to reduce wasted energy in a university campus building in Singapore. The system learns end-user preferences via the analysis of historical power consumption footprints. Specifically, individual and collaborative preferences in the usage of lights are extracted from actual usage data and the recommendations help users to automatically balance between their personalized visual comfort and energy efficiency.

\subsubsection{Context-aware}
Context-aware recommendation systems analyze historical power consumption data in different contexts and develop rule-based recommendations to ensure that end-users' preferences fit into them. 
In \cite{Shigeyoshi2013} an advisory recommendation system is proposed, which (i) analyzes energy consumption data in different contexts; (ii) keeps a record of recently issued recommendations in order to avoid repetitive recommendations; and (iii) conducts a social experiment, in which 47 end-users use the recommendation and provide feedback on its effectiveness. The experimental evaluation shows that randomly selected recommendations and  recommendation abundance have negative effects on the users.

In \cite{Luo2017} a personalized recommendation system is proposed. The system employs service recommendation schemes to derive possible end-users' interests and requirements related to energy efficiency of electrical devices. Next, identifies possibilities for saving energy and issues tailored recommendations to promote energy saving. Simultaneously, data retrieval methods are deployed to draw code words from textual appliance advertisements in order to increase energy saving awareness among users. The developed scheme first applies an energy disaggregation technique using a generalized particle filtering to infer appliance-level patterns from the main power consumption. Then, employs various inference rules to model the end-user's profile and preferences. Finally, tailored recommendations ensure an energy efficiency behavior are triggered, and the similarity between the user profile and device profile is measured, to rank the appliance advertisement and  generate recommendations. 

In \cite{Wei2018} authors begin with the assumption that energy efficiency can be dramatically limited if end-users are considered as \enquote{immovable objects}. Based on this, an energy efficiency recommendation system using end-user location is introduced. Specifically, two kinds of recommendations are generated based on location data. They are defined as (i) move recommendations that advise the end-user to move from a space to another; and (ii) shift-schedule recommendations that notify the end-user to arrive/depart a set amount of time earlier/later.  

In \cite{SARDIANOS2020394} a context-aware Recommender System (RS) is implemented based on (i) collecting data from smart sensor and actuators describing end-users energy consumption habits; and (ii) evaluating the triggered recommendations. In this regard, a RS called REHAB-C is developed, which can not only generate tailored energy-efficiency suggestions but it can also postpone or remove them based in the actual data and further store end-users' preferences. Specifically, contextual data are analyzed continuously and the recommendations are generated using a rule-based algorithm. 

Authors in \cite{Garcia2017} opt for the development of a recommendation system that combines the merits of information and communication technology (ICT) and social analysis to improve end-users energy consumption behavior. To that end, a context-aware collaborative algorithm is deployed to generate tailored recommendations for end-users. The implemented recommendation system includes a real-time localization module along with a wireless sensor network that provide real-time data about end-users' activities. The user context combines location and activity, which adds two more dimensions to the original user-item rating matrix of collaborative filtering. The context aware recommendation lead to a more fine grained tracking of user consumption behavior and thus to better recommendations.

\subsubsection{Rasch-based}
The Rasch model is a psychometric model used for the analysis of user responses to questionnaires, which aims to find the balance between the respondent's behavior and the difficulty of implementing the selected response.
Rasch-based recommendation systems are based on conducting a Rasch-based analysis to measure latent traits of end-users that are related to their energy consumption behavior and preferences. On the other side, they conduct survey studies to model the satisfaction of end-users to specific energy-saving actions and then based on the Rasch-analysis they generate personalized recommendations.
The Rasch-based power usage recommender system proposed in \cite{Starke2015} provides its end-users with personalized energy saving recommendations. The actions are recommended using a Rasch profile of users' behavior. In their framework, authors provided tailored suggestions (that match their Rasch profile) to 196 end-users via an online recommender platform.
In \cite{Starke2017} a recommender system to promote end-users' behavioral change is presented. Authors carried out two large review studies, where personalized energy-efficiency recommendations are evaluated for their feasibility and applicability by the users. Specifically, 79 energy-efficiency actions are drawn using a Rasch-based profile analysis to help end-users (i) make easy actions; (ii) improve system support; and (iii) collect their feedback and rate choice satisfaction.  

\subsubsection{Probabilistic models}
The recommendation systems of this type are based on the analysis of power consumption data. They develop probabilistic relational models in order to predict end-users preferences and then generate appropriate recommendations.
In \cite{Li7093924} historical power consumption patterns are analyzed using a continuous Markov chain model, which is based on a time-series investigation and multi-objective programming model. Moving forward, personalized recommendations are generated to support the use renewable energy solutions that are deployed in work environment.

In \cite{Wei9001078} authors propose a recommendation system to reduce wasted energy of commercial buildings using human-in-the-loop. The building energy efficiency task is modeled as a Markov decision process. Then, deep reinforcement learning are investigated for learning energy efficiency recommendations and engaging end-users in energy efficiency behaviors. Consequently, the adopted system learns user action, using a high energy efficiency potentiality, and thus notifies the end-users of a commercial building with recommendations. After that, feedback from end-users is utilized to understand what are the best energy efficiency recommendations.

\begin{table} [!b]
\caption{A comparison of related RS frameworks and their properties.}
\label{RelWorks}
\begin{center}
\resizebox{\textwidth}{!}{
\begin{tabular}{lllll}
\hline
Work & \begin{tabular}{@{}c@{}}Recommendation \\ Technique \end{tabular}
& Recommendations & Explanations & Application scenario  \\ \hline
Bravo et al. \cite{Bravo2019} & Case based & Electricity price based
recommendations & No & Households \\ 
Pinto et al. \cite{Pinto2018} & Case based & Energy reduction levels from similar cases & No & Public  \\ 
Zhang et al. \cite{Zhang8412100} & Collaborative filtering & Energy consumption plans \& tariffs & No & Households \\ 
ZHENG et al. \cite{ZHENG2020117775} & Collaborative filtering & Appliance-level consumption rates & No & Households \\ 
ReViCEE \cite{KAR2019135} & Collaborative filtering & Light
levels & No & Households \\ 
Shigeyoshi et al. \cite{Shigeyoshi2013} & Context-aware & Contextual based energy saving actions  & No & Households \\
Luo et al. \cite{Luo2017} & Context-aware & Tailored recommendations \& text ads & No & Households \\ 
Wei et al. \cite{Wei2018} & Context-aware & Move and shift-schedule actions & No & 
\begin{tabular}{@{}l@{}}Commercial \\ buildings \end{tabular}\\ 
REHAB-C \cite{SARDIANOS2020394} & Context-aware & Micro-moment personalised saving actions & No & \begin{tabular}{@{}l@{}}Academic \\ buildings \end{tabular}\\ 
Garcia et al. \cite{Garcia2017} & 
\begin{tabular}{@{}l@{}}Context-aware \\ \& Collaborative filtering  \end{tabular}
& Taolired advices
on end-users' activities & No & Households \\ 
Starke et al. \cite{Starke2015} & Rasch-based & Rasch profile based
recommendations of  & No & Households \\ 
&  & end-users' behavior &  &  \\ 
Starke et al. \cite{Starke2017} & Rasch-based & Rasch profile
receommendations based on & No & Households \\ 
&  &  a social experiment &  &  \\ 
Li et al. \cite{Li7093924} & Probabilistic relational & Tailored
recommendations to support the use & No & Work spaces \\ 
&  &  renewable energy solutions &  &  \\ 
Wei at al. \cite{Wei9001078} & Probabilistic relational & Recommendations
learned from actions having  & No & Commercial  \\ 
&  & high energy efficiency potentiality &  & buildings \\ 
This Work & \begin{tabular}{@{}l@{}}Explanaible\\Context-aware\end{tabular} & Energy saving actions, facts and reasoning & Yes & \begin{tabular}{@{}l@{}}Academic\\buildings\end{tabular}  \\ 
\hline
\end{tabular}
}
\end{center}
\end{table}

\subsection{Explainable recommender systems}

Based on the aforementioned analysis of different types of recommender systems for energy efficiency, and the recent trend for explainable Artificial Intelligence solutions, it seems that Recommendation Systems in the field of energy efficiency still lack Explainability of Persuasion features. 
Due to the emergent need for explainability in the recommendations provided to the users in a variety of scenarios, recent surveys like \cite{zhang2018explainable} try to review the different approaches for setting the various research questions regarding the explainable recommendation. For example, \cite{gao2019explainable} proposed a deep explicit attentive multi-view learning model to model multi-level features for explanation or the work in \cite{balog2019transparent} that examined an approach for creating a set-based recommendation model for transparent and textual explanations of movie recommendations. Towards a knowledge-based method for creating explainable item recommendations, authors in \cite{catherine2017explainable} illustrate a method for leveraging external knowledge in the form of knowledge graphs when information from content and product/item reviews is not available to generate explanations. Interpretable models, are based on transparent processes for deciding the recommendation lists, so it is easier to generate proper explicit feature-level explanations to justify why the model recommended specific items \cite{zhang2014explicit}. In the context of graph-based models, authors in \cite{he2015trirank} introduced an algorithm that ranks the vertices of a tripartite graph provide explanations for the top-ranked aspects-target user-recommended item triplets. However, in the domain of energy efficiency and recommendations for energy related behaviors there are few works, that usually attempt to explain the rules behind issuing a recommendation, which worth mentioning. They both come from the energy provider and prosumer domain.

Authors in \cite{grimaldo2019user} propose a user-centric and visual analytics approach to the development of an interactive and explainable day-to-day forecasting and analysis of energy demand in local prosumer environments. It is also suggested that this will be supported by a behavioral analysis to allow the analysis of potential relationships between consumption patterns and the interaction of prosumers with energy analysis tools such as customer portals and recommendation systems. A mixture of explainable machine learning approaches, such as kNN and decision trees, is used for dynamic simulation and explorative data processing.

\section{Proposed Methodology}\label{sec_methodology}

\subsection{Data model and architecture}
\begin{figure}[!htb]
\centering
\includegraphics[width=0.8\textwidth]{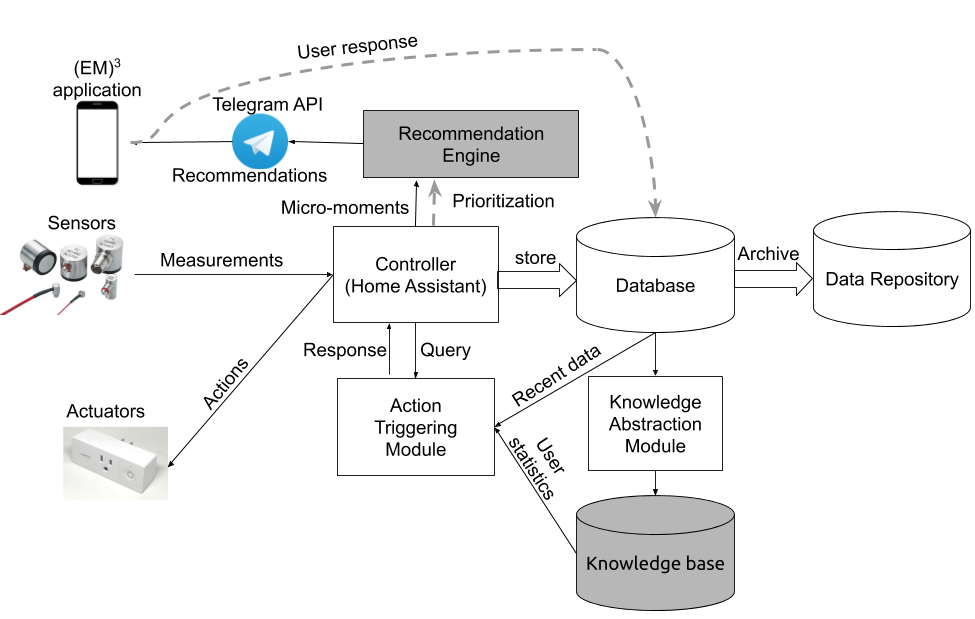} 
  \caption{The core architecture of the system and the explainable recommendation extensions (with grey color).}
\label{fig:architecture}
\end{figure}

The base for the explainable recommendations of this work, is the ecosystem depicted in Figure \ref{fig:architecture}, a combination of sensors, smart meters, actuators and orchestrating software that collaborate for promoting energy efficiency through smart on-time recommendations that build on the habit loop theory of behavioral change. Sensors (motion, temperature, humidity, light, etc.) capture user presence, and environmental conditions inside and outside of the monitored place, whereas smart meters capture electric power consumption per device thus creating a stream of measurement data that are stored in the platform database.

In order to efficiently handle the large amount of data generated from the sensors and smart meters an additional Knowledge Abstraction Module (KAM) has been developed. It comprises scripts, which process the sensor data streams and detect the respective micro-moments in real-time. Micro-moments are moments of special interest to the user, e.g. when the user exits the room, when the outside temperature (and humidity) conditions match user preferences, when a device has been used extensively, etc. The detected micro-moments, along with information about user preferences (e.g. the occupancy probability of each room at any moment of the day) are stored in the knowledge-base of the system.
The explainable recommendations framework of our system is supported by the appropriate data model, which organizes real-time data collected by the sensors and aggregated data that summarize recent device usage and room presence data (for a few weeks period) at 5-minute granularity. The former data are accumulated in the Database that stores detailed data for a few months period, whereas the latter are periodically updated to depict the recent user habits at any moment and are stored in what we refer to as Knowledge Base. Older sensor data are moved in a Data Repository for archiving purposes.

Real-time room occupancy, appliance consumption, and environment-related data along with knowledge about user habits are fed to the Action Triggering Module (ATM).
The ATM is developed as scripts in the open source home automation platform of Home Assistant \footnote{https://www.home-assistant.io/}. The platform allows custom scripting in Python and other languages and exposes several APIs for communicating with external applications and systems. 
The key channel for communicating recommendations to the end user is his/her smartphone and the Telegram application, which displays recommendations and collects user responses (i.e. accept, reject or ignore).

At the first step, the recommendation triggering step, the system periodically retrieves information about the user habits from the knowledge base and checks the recent sensor entries in the database in order to detect if it is the right micro-moment for a recommendation. 
For example, aggregated user presence data in a room are retrieved from the knowledge base and are used to compute the probability of room occupancy at any moment. This probability, along with recent motion sensor data from the database, and the recent user responses to related recommendations, are fed to the decision making algorithm that decides to send the recommendation.
Generated recommendations are displayed to the user's smartphone using the Telegram API, which is also installed on Home Assistant. This setup offers full flexibility to the messages sent to the end-user, which can be personalized to the user preferences, accompanied by explanations or additional persuasive facts, concerning the impact of an action.

What follows is the explanation of the recommendation. 
The current work focuses on the explainable recommendations, so the details of the Database schema and the architecture of the data collection system are given in \cite{varlamis2020bds} and are omitted.  
The emphasis of this work is on: i) the information stored in the Knowledge Base, explaining how this is exploited in order to provide explainable personalised recommendations; ii) the recommendation engine, explaining how user feedback is collected and processed in order to improve recommendation triggering. The additions to the core system architecture are depicted in grey dashed lines and the affected system modules are shaded grey.

\subsection{Recommendations for a purpose that come at the right moment}
One of the most important aspects of a real-time recommendation system is to be able to trigger an action recommendation at the correct moment for the user. Following the \textit{\textbf{micro-moments based recommendation}} strategy \cite{sardianos2019smartgreens}, the proposed system first detects micro-moments of special meaning to the daily user routine.
In terms of an energy efficiency recommendation system  \cite{alsalemi2019ieeesystems, alsalemi2020achieving, alsalemi_endorsing_2019}, this involves the identification of user's habitual actions, the analysis of the conditions that hold and the prediction of when actions will happen. For example, learn when the user turns the A/C on or off, in terms of time, environmental conditions, such as temperature and humidity (inside and outside).

Recommending the right energy saving action at the right moment can be very helpful for users who want to reduce their energy footprint 
\cite{sardianos2019smartgreens}. However, the chance to accept a recommendation increases when the \textbf{\textit{recommendation serves a purpose}}, and the purpose is clearly justified to the user. In the case of energy efficiency, the main purpose is to avoid the unnecessary usage of electrical and electronic devices (e.g. when contextual conditions allow it). An additional purpose that further reduces energy usage, can be to limit the usage of high energy-demanding devices.

In addition to the purpose of the recommendation, several facts that inform user on the benefits of an action can be beneficial for improving the recommendation acceptance. A \textbf{\textit{persuasive fact}} strengthens the recommendation, and helps user build a more energy efficient profile.

In this direction the proposed recommendation system is able to recognize the following aspects:
\begin {enumerate} 
\item the user \textbf{presence} in a room of interest, 
\item the general \textbf{context} --- which refers to the indoor and outdoor conditions i.e. temperature, humidity, luminosity, and 
\item user consumption \textbf{habits} in relation to various electrical and electronic appliances
\end {enumerate}. 
The system assists users to improve their energy footprint by recommending to turn off appliances when these are used without reason. 
This increases the probability of a recommendation to get accepted since it agrees to the user rational and the user intention to avoid unnecessary usage.
In order to clarify the intuition behind each recommendation, the system contains an explanation mechanism that creates a justification for each  recommended action, based on one of the aforementioned aspects.

In addition, the system provides users with persuasive facts, that are related to the savings that the user can have from a specific action or from adopting the new profile for a longer period. The facts, target the user incentives to change, and either promote an ecological profile or an economy profile depending on the user preferences.

The mechanisms that explain the recommendations and generate related facts are further analysed in the following.

\subsection{Explainable recommendations and human in the loop}

As mentioned before, in order to increase the user trust to the system and to maximize the recommendations' acceptance the system accompanies every recommendation for an energy saving action with: 
\begin {enumerate} 
\item a justification of \textbf{why} this action is recommended; and 
\item a fact that explains \textbf{what} would be the benefit for the user, if the recommendation is accepted.
\end {enumerate}

\begin{figure*}[!ht]
\centering
\includegraphics[scale=0.4]{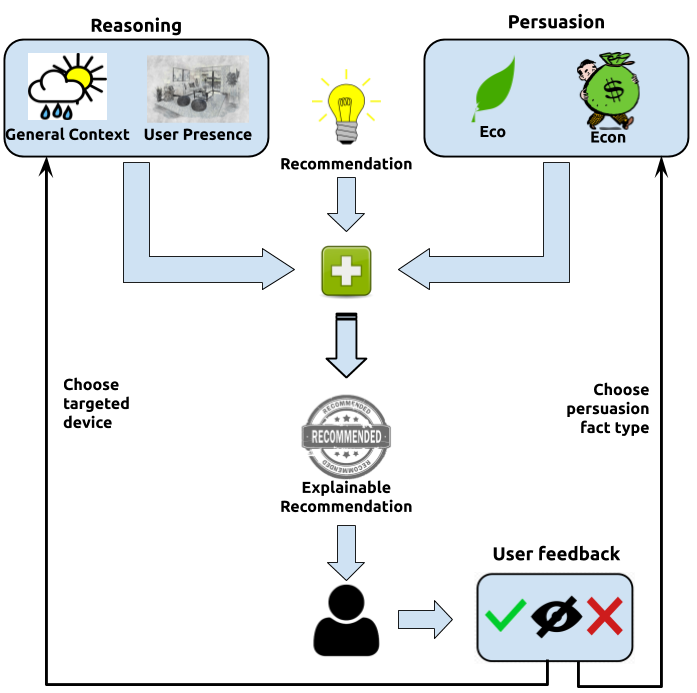}
  \caption{The flow of explainable recommendations.}
\label{fig:explRecommendations-methodology}
\end{figure*}

For supporting the above claims, work is done in two different aspects that define the two most essential characteristics that turn a recommendation for an energy saving action to an explainable recommendation as depicted in Figure \ref{fig:explRecommendations-methodology}:
\begin {enumerate}
\item \textbf{Reasoning:} This aspect considers the overall recommendation context and aim in providing detailed information on why the recommendation has been triggered. It can be information about the \textit{user status} (e.g. user is out of the room), about the \textit{appliance usage} (e.g. it is on for a long period) or the \textit{external conditions} that allow to turn off the appliance (e.g. outside temperature allows to open a window and turn off the air-conditioner).

\item \textbf{Persuasion:} This aspect builds on user preferences, incentives and long-term beliefs, and employs user feedback for choosing the right facts for each recommended action in order to make it more appealing for the user.
\end {enumerate}

The \textbf{\textit{Reasoning}} aspect of the explainable recommendation focuses on providing the reason(s) that triggered the recommendation. In an energy saving recommendation system, in which the main goal is to avoid the unnecessary usage of electrical and electronic devices, the reasons are tightly coupled with the excessive usage of devices. As a result, the reason behind a turn off action on a cooling or heating device can be that external environment conditions (e.g. temperature and humidity or simply the ``apparent temperature'') are similar to the inside ones and the device is still in use. Similarly, the reason behind a recommendation to turn off the room lights is that the natural light levels in the room are already high. Another reason is the unnecessary usage of some devices (e.g. cooling or heating, lights, monitors) when the user is out of the room. Although many of the above energy saving actions can be easily implemented using sensors and automations \cite{alsalemi2020achieving}, the use of recommendations brings human in the loop and allow him/her to decide on how to achieve the energy saving goals.

When the user manages to reduce the unnecessary usage of devices to the minimum, the next goal is to further reduce consumption by limiting the usage time of specific devices. The same recommendation explanation strategy is followed, but this time the reasons are related to time limits. For example, the recommendation to turn off the air conditioning comes a few minutes earlier than before. For this type of recommendations it is important to understand user habits (e.g. when or at what temperature the user turns the A/C on or off) \cite{sardianos2019smartgreens} in order to predict the next user movement (e.g. when the user is about to leave the room) \cite{sardianos2020model}. 

At this point, comes the second part of the explainable recommendation, which builds on the motivation behind user energy saving actions. These motivations affect user decision to turn off a device, and can either be ecological or economically driven. 
The \textbf{Eco} (a.k.a. Ecological) type of recommendations are targeted towards users that mostly focus on the environmental side of their energy consumption. These are users that are mainly motivated by contributing on the environmental movement by changing their own consumption habits and thus these recommendations focus on actions for reshaping the user's energy footprint with respect to the impact this will produce on the ecological aspect of the users' consumption habits. Examples of recommendation messages that fall in this category are e.g. "\textit{The total estimated number of kilowatt-hours from using the air-conditioning today is X, if you accept this recommendation you can contribute to a cleaner environment by reducing your energy consumption by Y\%!}". 
Such explanations aim to increase the persuasiveness of the recommendation and to be an arousal factor to the users that are interested in having a good ecological behavior, but need a trigger to motivate them.
The \textbf{Econ} (a.k.a. Economical) type of recommendations have been employed to target the users that prioritize their financial savings over ecology. It is an alternative to the Eco type of recommendation messages for users that are mostly concerned about the amount of money they tend to spend every month for energy consumption (either electricity or gas) to cover their personal consumption needs.

\subsection{Personalising the explainable recommendations}

As it is depicted in Figure \ref{fig:explRecommendations-methodology}, the life-cycle of an explainable recommendation does not end with its delivery to the user. The user response to each recommendation is valuable for the system and is accounted when issuing the next recommendation. Even if the user decides to ignore the recommendation and to not interact with the system, the response is recorded and the next recommendation is adjusted accordingly.

More specifically, in terms of the persuasion mode, the system begins with giving equal importance to the Eco and Econ profiles and the associated persuasion fact is selected with equal probability from the pool of Eco and Econ persuasion facts related to the targeted device. The positive (or negative) response of the user to the recommended action is counted in favor (or against) the type of the persuasion fact. For example, if the user gets a recommendation with an ecology-related fact and decides to reject it, his/her Eco profile is penalised. Similarly, the Eco profile gets a bonus when the user accepts the recommendation. If the user ignores the recommendation, assuming that he/she was not aware of the recommendation and the explanation, we neither bonus nor penalise his/her profile.

On a different line, the user response to a recommendation that interacts with a device is used as feedback for the mechanism that triggers the recommendation. For example, when a user decides to accept a recommendation to turn off the air cooling device, which was triggered because the external environment temperature drops in the evening, the successful recommendation is recorded in order to be repeated in the first opportunity. When a large number of acceptances is recorded for a recommendation, the recommendation can be marked for automatic acceptance (i.e. an automation) in the future. In the event of a rejected recommendation, the information is recorded and the recommendation engine can set a temporary pause to the recommendations for the specific device. When the user keeps rejecting a recommendation, it will be permanently paused automatically, or after the user approval. When the user ignores a recommendation, the system temporarily pauses this recommendation for a few minutes. 

Based on the above analysis, all the user and environment data are recorded and processed in order to issue a recommendation. The user responses (or no response) is also recorded and processed in order to update the user preferences profile. In the subsection that follows, we detail on the data model of the proposed approach and the architecture employed for issuing explainable recommendations.

\section{Experimental Evaluation}
\label{sec_evaluation}

\subsection{Experimental setup}
The most important criterion in the evaluation of a recommendation system is the \textbf{\textit{acceptance rate}} of the recommendations it creates. When the system supports a predefined objective, such as improving energy efficiency, then we can also evaluate the recommendation system based on whether it helps users to reach their objective.
The effectiveness of the explainable recommendations is an important measure for identifying whether the recommendation engine achieves its goal or not. In order to evaluate the acceptance of the personalised recommendations and the effect of persuasion and explanations to this acceptance, we performed a study on a group of users. 
The group of users comprises office users, who all evaluated the same scenario. The reason behind this, is to avoid the individual user bias and better understand the effect of explainable recommendations to users' decisions. For this purpose, all users were exposed to the same simulated scenario, that was based on the actual days in the office of another user. 

The data used for the evaluation of our scenario have been collected for several consecutive days from the installation of that users' office in the facilities of the university. Based on the data collected from this office, and the surrounding (outdoor) conditions, the same energy saving action recommendations were created and presented to all the participants of the study.

To be more precise, the sensor data used in the simulated scenario were actually collected from the real sensor setup in the office facilities during consecutive days of office use and are used as a starting baseline in order to identify the environmental conditions and user context and start presenting recommendations to the users. Once each user starts to receive recommended energy actions and responds to these recommendations, the system adjusts to each user's preferences. This means that all the users who participated in the simulation used the same set of sensor data as if they were actually in this office during the period of data collection.

Although, all participants run the same scenario (e.g. the outdoor temperature and humidity conditions change similarly for all users, user presence in the room is the same for all, etc.) the decisions of the users affect the conditions inside the room. More specifically, during the evaluation process, recommendations are triggered and displayed to all users at the same time and users have the option to accept, ignore or reject them. Based on the decision of each user (e.g. to turn off the lights or the A/C as recommended) the indoor conditions change accordingly. Given that external conditions change in a similar way for all users, a varying number of recommendations can be issued to each user during the scenario execution.

Despite the current experimental setup, in which our problem setting mainly focused on monitoring and controlling the office lights, the A/C unit and the PC monitors as a proof of concept, the scalability of our proposed framework in larger cases is one of the principles of our architecture. Since each appliance is managed autonomously or in combination with other devices based on the user's goals and automations the system can flawlessly scale to larger case scenarios with more devices and actions needed. In addition, the requirements needed for the framework to run are not resource demanding and do not depend on the number of monitored appliances, so it is easy to reproduce this setup in larger spaces.

Since the acceptance of recommendations can be boosted by providing additional \textbf{\textit{persuasive facts}} to the user, we include the following type of facts:
\begin{itemize}
    \item The Eco type of persuasive facts which build on the ecological impact of the energy consumption.
    \item The Econ type of persuasive facts which promote the economic impact of the energy consumption to the user.
\end{itemize}

Recommending the correct action at the correct moment is a critical aspect that affects the recommendations' performance. It is also important to provide the user with information about the reasons that triggered each recommendation.
For the purpose of the evaluation process, we focus only on recommendations about turning-off the A/C unit and the office lights. The \textbf{\textit{reasons}} for triggering a recommendation can mainly be divided into two types:
\begin{enumerate}
  \item Recommendations are  triggered because the user has left the room and left an appliance (A/C or lights) on, thus consuming energy without reason, and
  \item Recommendations are triggered while the user is still in the room and has a device at the on state, but the outdoor conditions (light or temperature) allow to avoid excessive or unreasonable usage, e.g. by opening a window to cool the room or allow natural light.
\end{enumerate}

\begin{figure}[!ht]
\begin{subfigure}[t]{.33\textwidth}
    \centering
    \includegraphics[trim=0 45cm 0 0,clip,width=1.0\columnwidth]{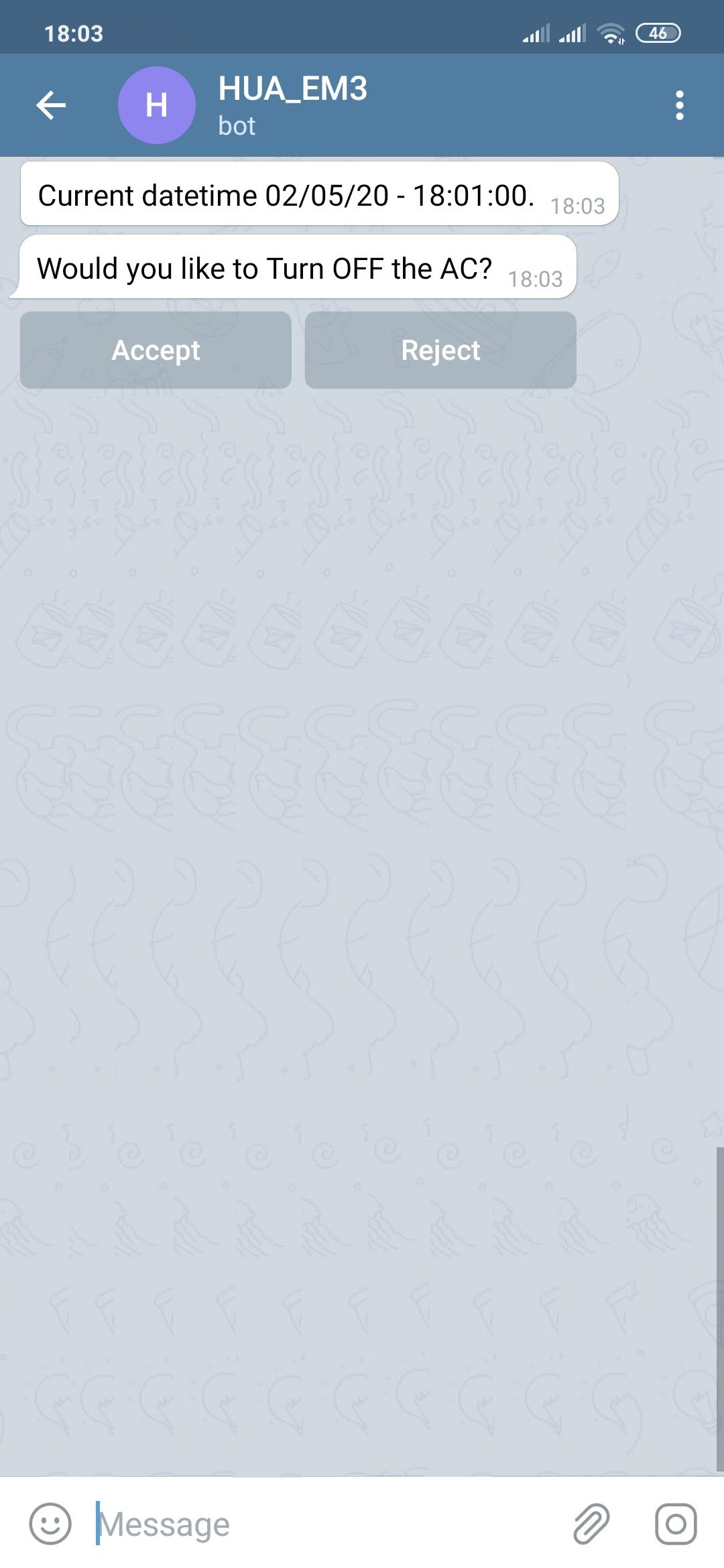}
    \caption{}
    \label{fig:plain-recommendation}
\end{subfigure}
\begin{subfigure}[t]{.33\textwidth}
    \centering
    \includegraphics[trim=0 41cm 0 0,clip,width=1.0\columnwidth]{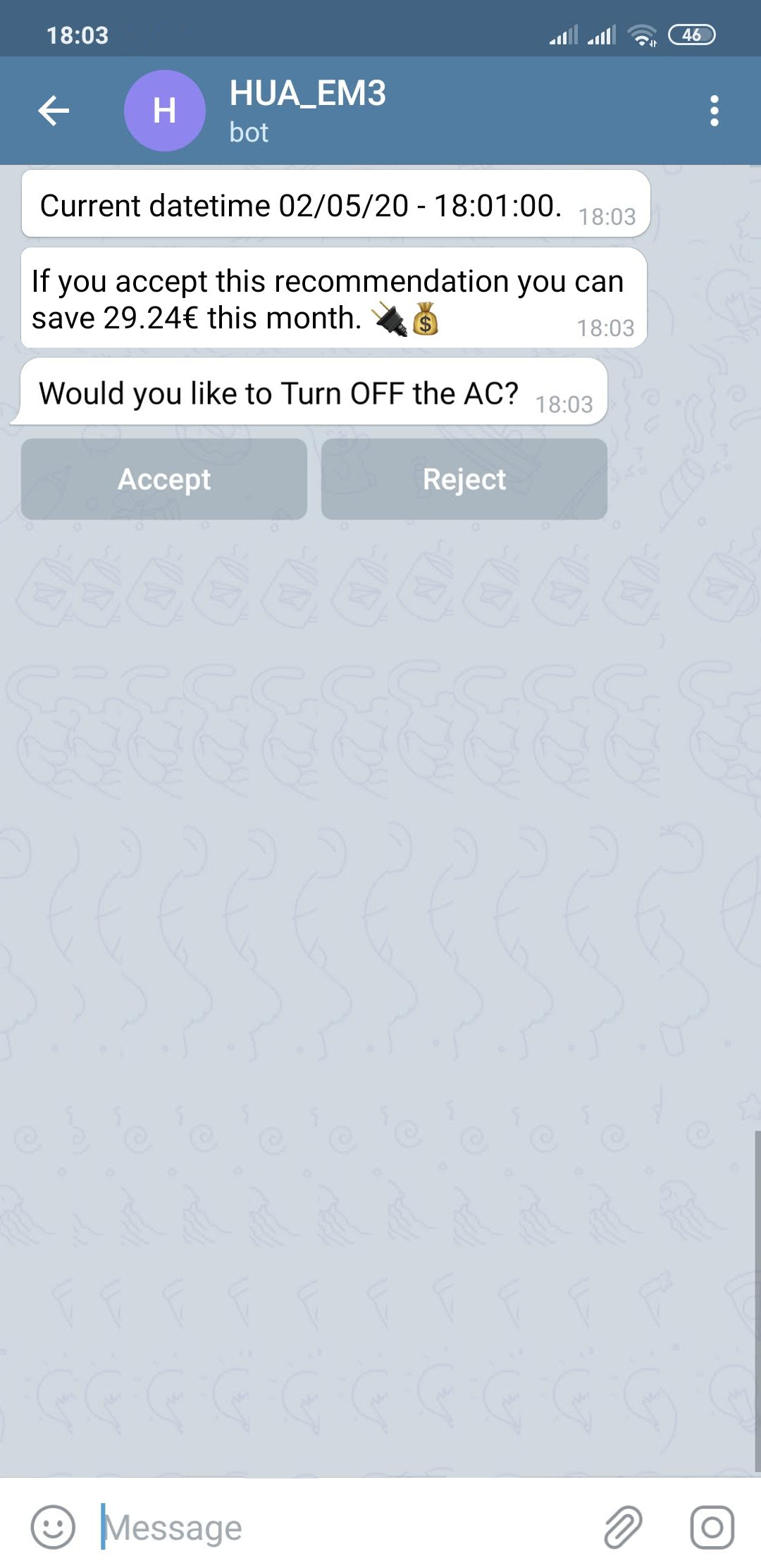}
    \caption{}
    \label{fig:persuasive-recommendation}
\end{subfigure}
\begin{subfigure}[t]{.33\textwidth}
    \centering
\includegraphics[trim=0 45cm 0 0,clip,width=1.0\textwidth]{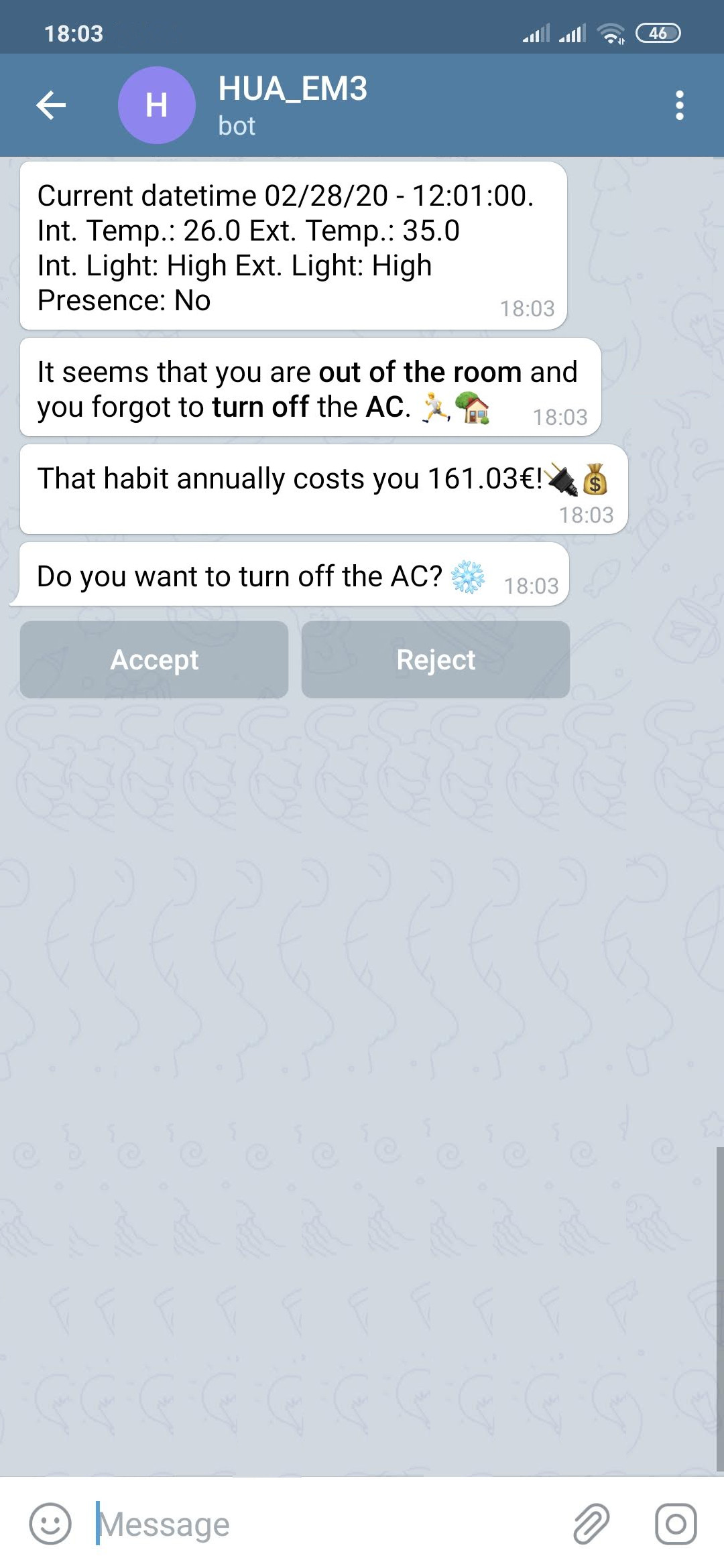}
    \caption{}
    \label{fig:explainable-recommendation}
\end{subfigure}
\caption{Examples of: (a) plain recommendations in \textit{\textbf{Scenario I}}, (b) recommendations with persuasive facts for \textit{\textbf{Scenario II}}, and (c) recommendations accompanied with the reasoning part to form the explainable recommendations for \textit{\textbf{Scenario III}}.}
\label{fig:recommendations-per-scenario}
\end{figure}

The experimental setup evaluates three versions of the ``week in the smart office'' scenario, which is explained in more details in the following subsection. The key point that differentiates the three versions lies in the content of the messages that were delivered to the end-users as well as the persuasive factors and the explanations each recommendation included in its body, and can be summarized in the following:
\begin{enumerate}
  \item The simple version recommendations include no particular informative content apart from the date and time that the recommendation has been created, and the prompt for the recommended action (Figure \ref{fig:recommendations-per-scenario}(a)).
  \item In the second variation of the experiment, each recommendation includes
  a customized persuasive fact that describes the expected impact of the user's consumption habit (Figure \ref{fig:recommendations-per-scenario}(b)).
  \item Finally, in the full version, the explainable and persuasive recommendations also include information about the indoor and outdoor conditions and the user presence, as well as a message that informs the user for the reason that triggered the recommendation (Figure \ref{fig:recommendations-per-scenario}(c)).
\end{enumerate}

Since the decisions of users to the same recommendation can differ, the sequence of recommendations also differs per user, thus giving useful insights about the type of recommendations that each user tends to accept or reject. The same variability occurs to the type of persuasive facts that accompanies each recommendation. More details about the base scenario that was the basis for all users, and the way recommendations are experimental setup and the evaluation process are provided in the subsections that follow.
In total 8 users participated in this evaluation process, which revealed many interesting finding concerning user preferences and the impact of explanations and persuasive facts to the recommendation acceptance.

\subsection{The scenario: A typical week in the office}

Our system has already been deployed in several offices of our University's facilities, as described in Section \ref{sec_methodology} and in \cite{alsalemi2019ieeesystems, sardianos2019smartgreens} using a set of various sensors that record data about device usage, consumption (e.g. energy consumption per device), user presence and the general context (e.g. interior and exterior temperature, humidity, and luminosity). Energy consumption data along with occupancy information and contextual factors derived from the sensor data are used to extract the users’ consumption habits. More specifically the analysis of sensor data, allows the system to identify user consumption habits, which are represented as device usage or room presence patterns (e.g. turn on the A/C at 17:00) and the contextual conditions under each usage is performed (e.g. when the outdoor temperature is above 33$^{\circ}$ for example).

On the basis of the learned user patterns it is possible to trigger proper energy saving action recommendations when certain conditions are met (e.g. when the outdoor temperature is lower than indoor so turning off the A/C and opening a window may be a more suitable alternative).

For the experimental evaluation and the creation of the evaluation simulation scenario, data from a single user office for one week has been collected. The focus was on user presence, and the environmental conditions (the luminosity levels and the temperature) inside and outside the room.

\noindent\textbf{Environmental conditions:}
In order to capture the contextual information of the scenario (e.g. the environmental conditions of the office), we employed two DHT-22 temperature and humidity sensors with an operating range of -40-80$^{\circ}$C and 0-100\% for ambient temperature and relative humidity respectively, and the sensor measurements recorded during a three months period was used for the evaluation scenario.

\noindent\textbf{Room occupancy:}
A key aspect on identifying excessive energy consumption is the ability to recognize when there is no user in the room that an energy consuming device is still turned on. In the office setup used for this evaluation, two HC-SR501 motion sensor units were used in order to record room occupancy which is also stored in the system's data store. The outcomes of the initial data analysis process over the room occupancy data, a set of timeslots where the user has been identified to be absent from the office, resulting in the weekly compiled list of room occupancy slots that is depicted in Table \ref{table:occupancy-time-slots}.

\begin{table}
\caption{The detected office occupancy hours, using our system's setup}
\label{table:occupancy-time-slots}
\centering
\arrayrulecolor{black}
\resizebox{0.7\columnwidth}{!}{
\begin{tabular}{|l|l|l|l|l|l|} 
\hline
\diagbox{Time of Day}{Day of Week}        & \multicolumn{1}{c|}{{\cellcolor[rgb]{0.9,0.9,0.9}}Monday} & \multicolumn{1}{c|}{{\cellcolor[rgb]{0.9,0.9,0.9}}Tuesday} & \multicolumn{1}{c|}{{\cellcolor[rgb]{0.9,0.9,0.9}}Wednesday} & \multicolumn{1}{c|}{{\cellcolor[rgb]{0.9,0.9,0.9}}Thursday} & {\cellcolor[rgb]{0.9,0.9,0.9}}Friday  \\ 
\hline
\rowcolor[rgb]{0,0,0} {\cellcolor[rgb]{0.9,0.9,0.9}}8AM - 9AM    &                                                             & {\cellcolor[rgb]{1,1,1}}                         &                                                                &                                                               &                                         \\ 
\hhline{|->{\arrayrulecolor[rgb]{0,0,0}}---->{\arrayrulecolor[rgb]{1,1,1}}->{\arrayrulecolor{black}}|}
\rowcolor[rgb]{0,0,0} {\cellcolor[rgb]{0.9,0.9,0.9}}9AM -10AM    &                                                             &                                                              &                                                                &                                                               & {\cellcolor[rgb]{1,1,1}}    \\ 
\hhline{|->{\arrayrulecolor[rgb]{0,0,0}}--->{\arrayrulecolor[rgb]{1,1,1}}-->{\arrayrulecolor{black}}|}
\rowcolor[rgb]{0,0,0} {\cellcolor[rgb]{0.9,0.9,0.9}}10AM - 11AM  &                                                             &                                                              &                                                                & {\cellcolor[rgb]{1,1,1}}                          & {\cellcolor[rgb]{1,1,1}}    \\ 
\hhline{|->{\arrayrulecolor[rgb]{0,0,0}}-->{\arrayrulecolor[rgb]{1,1,1}}--->{\arrayrulecolor{black}}|}
\rowcolor[rgb]{1,1,1} {\cellcolor[rgb]{0.9,0.9,0.9}}11AM - 12PM & {\cellcolor[rgb]{0,0,0}}                        & {\cellcolor[rgb]{0,0,0}}                         &                                                                &                                                               &                                         \\ 
\hhline{|->{\arrayrulecolor[rgb]{1,1,1}}----->{\arrayrulecolor{black}}|}
\rowcolor[rgb]{1,1,1} {\cellcolor[rgb]{0.9,0.9,0.9}}12PM - 1PM   &                                                             &                                                              &                                                                &                                                               &                                         \\ 
\hhline{|->{\arrayrulecolor[rgb]{1,1,1}}--->{\arrayrulecolor[rgb]{0,0,0}}->{\arrayrulecolor[rgb]{1,1,1}}->{\arrayrulecolor{black}}|}
\rowcolor[rgb]{1,1,1} {\cellcolor[rgb]{0.9,0.9,0.9}}1PM - 2PM    &                                                             &                                                              &                                                                & {\cellcolor[rgb]{0,0,0}}                          &                                         \\ 
\hhline{|->{\arrayrulecolor[rgb]{0,0,0}}->{\arrayrulecolor[rgb]{1,1,1}}->{\arrayrulecolor[rgb]{0,0,0}}-->{\arrayrulecolor[rgb]{1,1,1}}->{\arrayrulecolor{black}}|}
\rowcolor[rgb]{0,0,0} {\cellcolor[rgb]{0.9,0.9,0.9}}2PM - 3PM    &                                                             & {\cellcolor[rgb]{1,1,1}}                         &                                                                &                                                               & {\cellcolor[rgb]{1,1,1}}    \\ 
\hhline{|->{\arrayrulecolor[rgb]{0,0,0}}--->{\arrayrulecolor[rgb]{1,1,1}}->{\arrayrulecolor[rgb]{0,0,0}}->{\arrayrulecolor{black}}|}
\rowcolor[rgb]{0,0,0} {\cellcolor[rgb]{0.9,0.9,0.9}}3PM - 4PM    &                                                             &                                                              &                                                                & {\cellcolor[rgb]{1,1,1}}                          &                                         \\ 
\hhline{|->{\arrayrulecolor[rgb]{0,0,0}}-->{\arrayrulecolor[rgb]{1,1,1}}->{\arrayrulecolor[rgb]{0,0,0}}-->{\arrayrulecolor{black}}|}
\rowcolor[rgb]{0,0,0} {\cellcolor[rgb]{0.9,0.9,0.9}}4PM - 5PM    &                                                             &                                                              & {\cellcolor[rgb]{1,1,1}}                           &                                                               &                                         \\ 
\hhline{|->{\arrayrulecolor[rgb]{1,1,1}}->{\arrayrulecolor[rgb]{0,0,0}}->{\arrayrulecolor[rgb]{1,1,1}}->{\arrayrulecolor[rgb]{0,0,0}}-->{\arrayrulecolor{black}}|}
\rowcolor[rgb]{0,0,0} {\cellcolor[rgb]{0.9,0.9,0.9}}5PM - 6PM    & {\cellcolor[rgb]{1,1,1}}                        &                                                              & {\cellcolor[rgb]{1,1,1}}                           &                                                               &                                         \\ 
\hhline{|->{\arrayrulecolor[rgb]{0,0,0}}->{\arrayrulecolor[rgb]{1,1,1}}--->{\arrayrulecolor[rgb]{0,0,0}}->{\arrayrulecolor{black}}|}
\rowcolor[rgb]{1,1,1} {\cellcolor[rgb]{0.9,0.9,0.9}}6PM - 7PM    & {\cellcolor[rgb]{0,0,0}}                        &                                                              &                                                                &                                                               & {\cellcolor[rgb]{0,0,0}}    \\ 
\hhline{|->{\arrayrulecolor[rgb]{0,0,0}}->{\arrayrulecolor[rgb]{1,1,1}}->{\arrayrulecolor[rgb]{0,0,0}}-->{\arrayrulecolor[rgb]{1,1,1}}->{\arrayrulecolor{black}}|}
\rowcolor[rgb]{0,0,0} {\cellcolor[rgb]{0.9,0.9,0.9}}7PM - 8PM    &                                                             & {\cellcolor[rgb]{1,1,1}}                         &                                                                &                                                               & {\cellcolor[rgb]{1,1,1}}    \\ 
\hhline{|->{\arrayrulecolor[rgb]{1,1,1}}->{\arrayrulecolor[rgb]{0,0,0}}->{\arrayrulecolor[rgb]{1,1,1}}->{\arrayrulecolor[rgb]{0,0,0}}->{\arrayrulecolor[rgb]{1,1,1}}->{\arrayrulecolor{black}}|}
\rowcolor[rgb]{1,1,1} {\cellcolor[rgb]{0.9,0.9,0.9}}8PM - 9PM    &                                                             & {\cellcolor[rgb]{0,0,0}}                         &                                                                & {\cellcolor[rgb]{0,0,0}}                          &                                         \\ 
\hhline{|->{\arrayrulecolor[rgb]{1,1,1}}--->{\arrayrulecolor[rgb]{0,0,0}}->{\arrayrulecolor[rgb]{1,1,1}}->{\arrayrulecolor{black}}|}
\rowcolor[rgb]{1,1,1} {\cellcolor[rgb]{0.9,0.9,0.9}}9PM - 10PM   &                                                             &                                                              &                                                                & {\cellcolor[rgb]{0,0,0}}                          &                                         \\
\hline
\multicolumn{1}{l}{}                                                           & \multicolumn{1}{l}{}                                        & \multicolumn{1}{l}{}                                         & \multicolumn{1}{l}{}                                           & \multicolumn{1}{l}{}                                          & \multicolumn{1}{l}{}                    \\
\multicolumn{1}{l}{}                                                           & \multicolumn{1}{l}{{\cellcolor[rgb]{0,0,0}}}    & \multicolumn{4}{l}{User presence was identified in the office}\\
                                                                     
\multicolumn{1}{l}{}                                                           & \multicolumn{1}{|c|}{{\cellcolor[rgb]{1,1,1}}}    & \multicolumn{4}{l}{User was absent from the office}  \\
\cline{2-2}
\end{tabular}
}
\end{table}

\noindent\textbf{Consumption habits:}
Based on the consumption data analysis, the system identifies the energy consumption preferences of the user in terms of when the user tends to turn-on and off certain devices and combining this info with the contextual information recorded from the indoor and outdoor sensors, the system "knows" when and under what conditions the user's energy consumption habits tend to occur (e.g. the luminosity levels of the room that triggers the user to switch on the lights).

\subsection{Recommendations delivery and user feedback}
\label{ss_delivery}

\begin{figure}[!ht]
\begin{subfigure}[t]{.33\textwidth}
    \centering
    \includegraphics[trim=0 45cm 0 0,clip,width=1.0\textwidth]{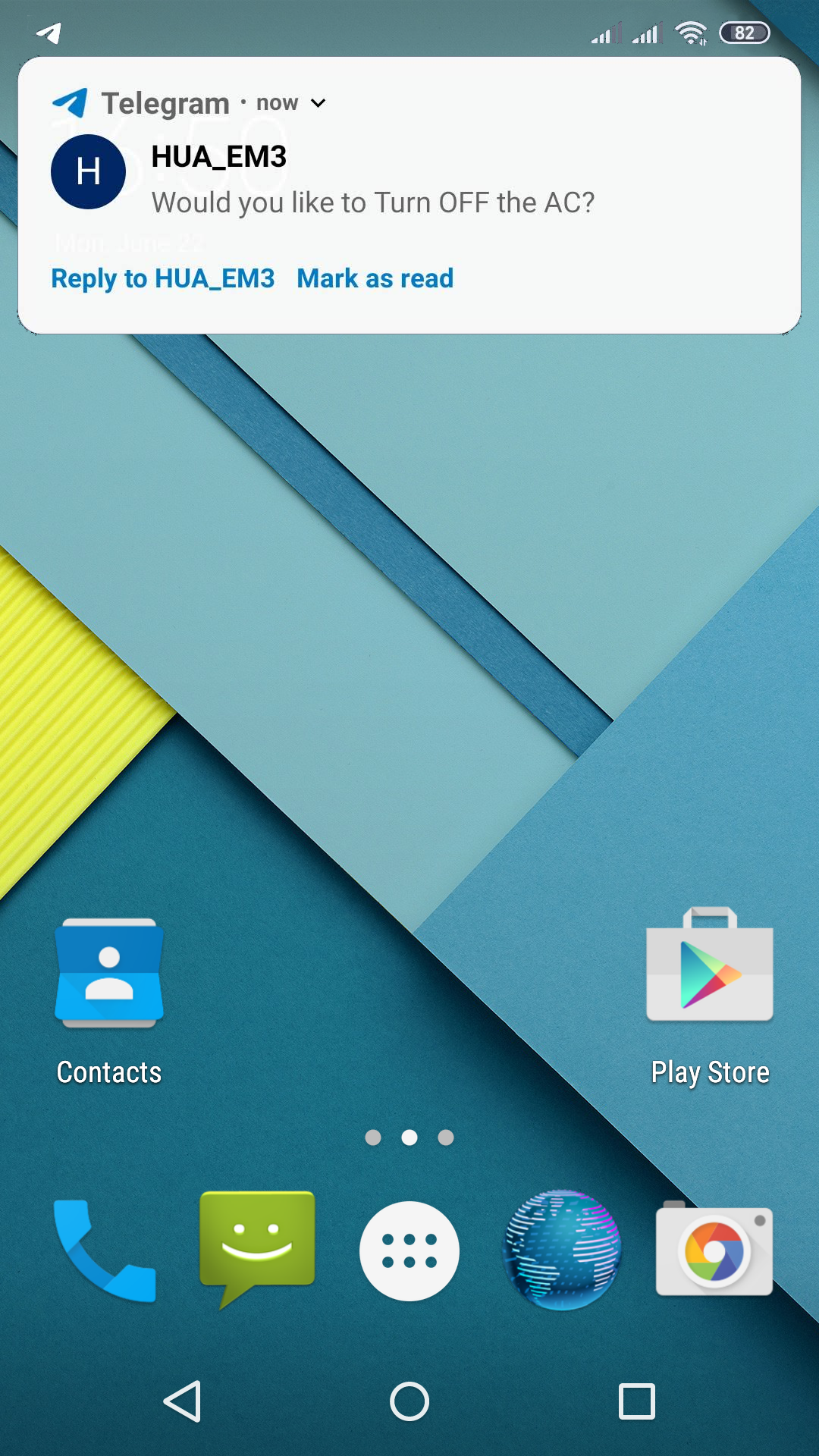}
    \caption{}
\end{subfigure}
\begin{subfigure}[t]{.33\textwidth}
    \centering
    \includegraphics[trim=0 60cm 0 0,clip,width=1.0\columnwidth]{recommendation-ac.jpg}
    \caption{}
\end{subfigure}
\begin{subfigure}[t]{.33\textwidth}
    \centering
    \includegraphics[trim=0 60cm 0 0,clip,width=1.0\textwidth]{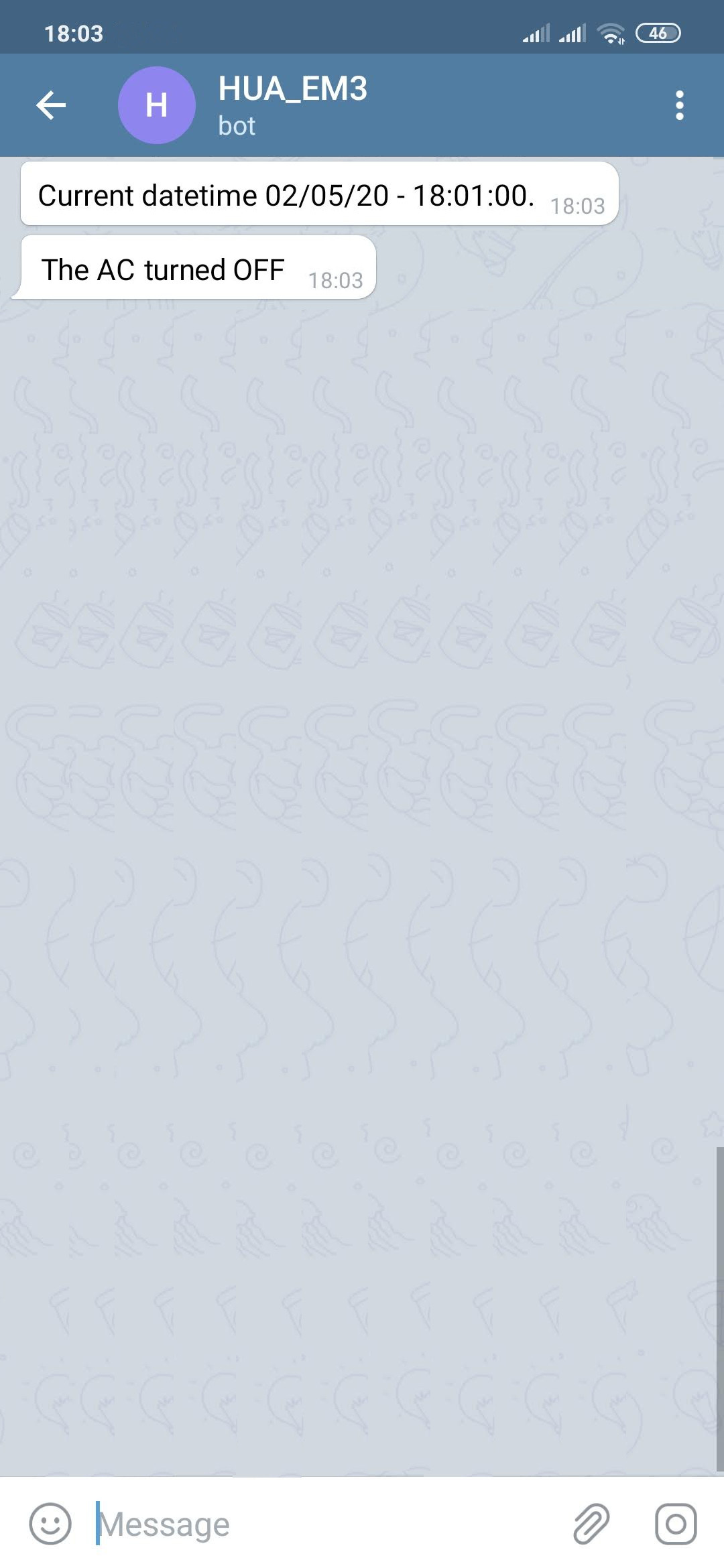}
    \caption{}
\end{subfigure}
\caption{From left to right: (a) an example of a pop up recommendation presented to the user's smartphone, (b) the plain recommendation without any persuasive fact or explanation and the user options, and (c) the result after a user accept response.}
\label{fig:recommendations-presentation}
\end{figure}

When the proper conditions occur, the system generates a recommendation that is presented to the user's smartphone as a pop up notification, like the one presented in Figure \ref{fig:recommendations-presentation}. In the plain recommendation scenario, when the user opens the recommendation in the Telegram app, the recommendation message provides only a timestamp (i.e. the date and time when the recommendation occurred) and the recommended action (e.g. turn off the A/C), in order to allow the evaluators to put themselves into the context under which this recommendation was generated. After the current datetime, the phrasing of the recommended action follows which is accompanied by two ``Accept" and ``Reject" buttons, that allow the user to respond to this recommendation (Figure \ref{fig:recommendations-presentation}).

Upon acceptance of a recommended action, the system automatically sends a turn-off signal to the respective device and an acknowledgement to the Telegram message that informs the user that this action has been fulfilled (Figure \ref{fig:recommendations-presentation}(c)). If the user decides to reject the recommendation, the system takes into consideration the user's negative response to this recommended action and adjusts the way this recommendation has to be presented again in the future.

Every time that a recommendation is delivered to the user, he/she has a pending time (20 seconds) for accepting or rejecting the recommendation, else the recommendation is considered as ignored. Depending on each user's answer the sequence and timing of recommendations is different for each user. The experiment runs in simulated time and the process is sped up, when no recommendations are pending for any user.

In particular, when a turn-off action gets rejected (i.e. the user prefers to leave the appliance turned on), the recommendation engine does not reissue this recommendation for a period of one hour and if and only if the conditions that triggered this recommendation (e.g. if the user is still absent from the office, etc.) are still valid. In the case that a recommendation has been ignored or simply the user failed to respond during the allowed time window to a recommendation, then this recommended action is held for a period of 10 minutes and then is sent again to the user, once again given that the trigger conditions are still met. The system will continue to issue the same recommendation for a maximum time window of one hour when the recommendation is ceased and is not displayed again to the end-user.

With the above configuration, the same ``a week in the office'' scenario may result to a different sequence of recommendation messages for the users. This leads to a total time of 60-70 minutes needed for running one scenario for one week (5 working days) for all the users. The same experiment is repeated two more times using persuasive and explainable recommendation messages as explained in the following.

\subsection{Explainable recommendation messages}
For Scenarios II and III, that both employ persuasive facts to accompany the recommended action to increase the possibility to accept a recommendation, the proposed system personalizes the messages addressed to the user. 
Also, in Scenario III, the explainable recommendations produced by the recommendation engine fit to the recommendation context and provide a verbal description of the conditions that triggered the recommendation. 

\begin{figure}[!hbt]
\centering
\begin{subfigure}[b]{.30\textwidth}
  \centering
  \includegraphics[trim=0 40cm 0 0,clip,width=1.0\columnwidth]{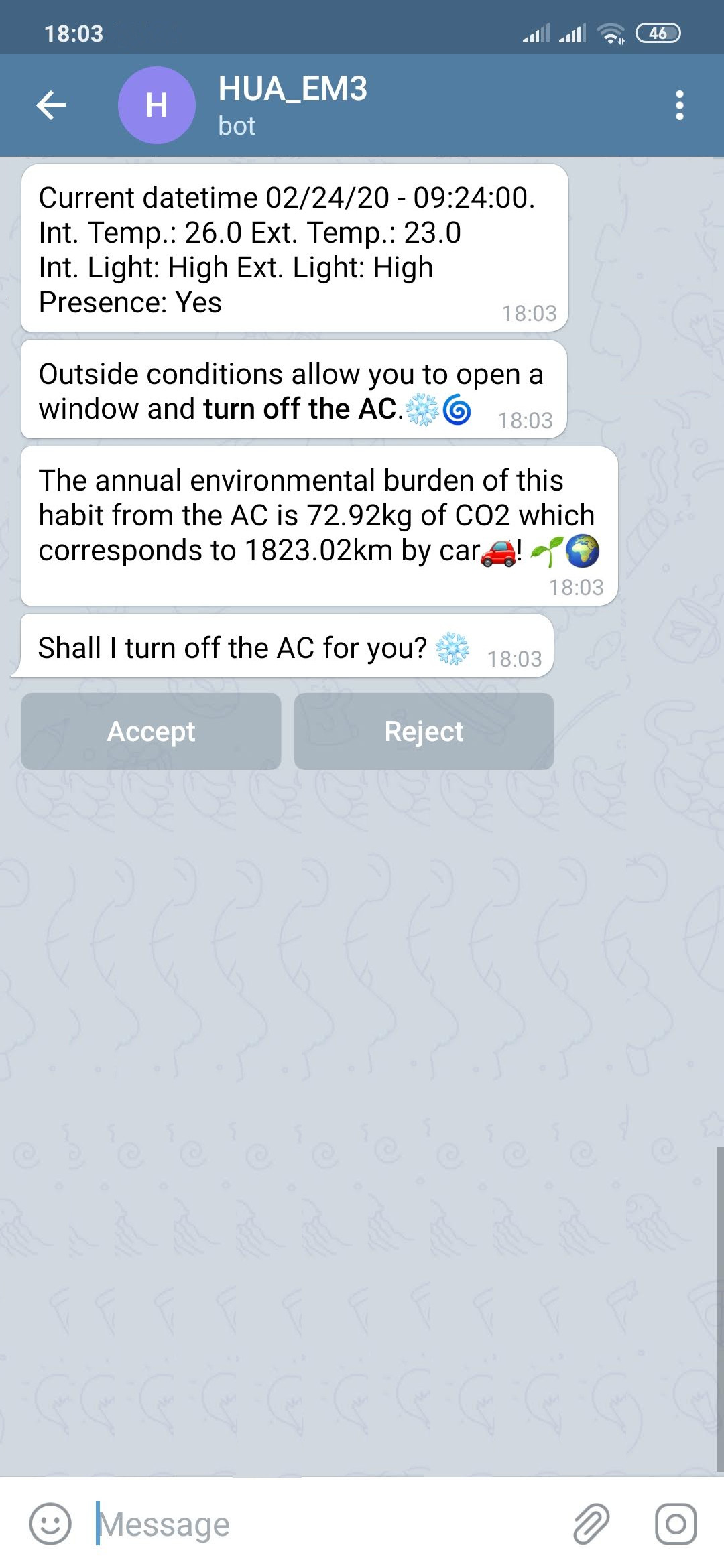}
  \label{fig:eco-recommendation}
\end{subfigure}
\begin{subfigure}[b]{.30\textwidth}
  \centering
  \includegraphics[trim=0 40cm 0 0,clip,width=1.0\columnwidth]{recommendation-ac-2-econ-explainable.jpg}
  \label{fig:econ-recommendation}
\end{subfigure}
\caption{Examples of personalized explainable recommendations presented to the user accompanied by the reasoning and the persuasive facts.}
\label{fig:types-of-recommendation}
\end{figure}

Two examples of explainable recommendations are depicted in Figure \ref{fig:types-of-recommendation}. Recommendations are composed by two different sections, which serve the explainable and persuasive recommendations scenario respectively. The messages contain visual cues that facilitate users to understand the reason of each recommendation. The first message provides the date and time information as in the plain recommendation scenario, as well as information about: i) the internal and external temperature, ii) the internal and external light levels, iii) the user presence in the room. The second message contains a verbal explanation of the reason that triggered the recommendation, which summarizes the values presented in the first message. The third message contains the persuasive fact, which is Eco or Econ and contains an estimation for the actual saving (e.g. by turning off the A/C earlier than usual) or a projection of this saving in month and yearly level.  
This enables all the different evaluators that participate in this evaluation process to put themselves into the context that triggered this recommendation and thus making this process as realistic as possible. 

The persuasive part of the recommendation (third message) appears in both Scenarios II and III, whereas the explanation part (first and second message) appears only in Scenario III.

In the context of why a recommendation has been triggered, the \textbf{\textit{explanations}} produced for the users are categorized into:
\begin{itemize}
    \item The user is \textbf{\textit{out the office}} while the devices are still on, and
    \item The user is still in the office, with the devices switched on, but the indoor/outdoor conditions allow the \textbf{use of alternate methods} for achieving similar results (e.g. use natural lighting instead of office lights to light up the office, open the window to cool down the room instead of turning on the A/C, etc.).
\end{itemize}

In the examples of Figure \ref{fig:types-of-recommendation} the recommendation on the left was created because the indoor and outdoor context (e.g. the difference in the indoor and outdoor temperature) have been identified as suitable for turning off the A/C and opening the window to achieve the same results. On the contrary, the one on the right notifies the user that the system has detected the user absence from the room and the A/C was on. This part of recommendation plays an important role on engaging users to accept the recommended actions since the explainability aspect of the personalized recommendations makes it easier for the users to understand the ``flaws" in their consumption habits and thus increase their trust in the system that assists them to change their energy profile.

The \textbf{\textit{persuasion facts}} act as supplements to the main explainability of the recommendation and try to persuade the users to accept the recommendations by pointing out the benefits for the user when accepting the recommended actions.
The personalization of this facts is based on detecting whether the user values the ecological factor higher than the economical one or vice versa based on the acceptance or rejection of the respective recommendations issued at any stage of the experiment. Based on this, it adjusts the type of facts that comes with the recommendations that follow, to match the user's preference. So, when a new recommendation is generated, the system calculates the consumption of the targeted device from the time this device was turned on until that moment and makes a projection of the total $CO_2$ emissions and the total cost in \EUR (or other currencies) from the usage of the device for this specific period. The fact is then enriched with the total energy consumption, and the total energy savings (in the case of Eco recommendations) or money savings (in the case of Econ recommendations), and is presented along with the recommended action to the user. The resulting recommendation is more informative than the previous one (that contains only time information) and is expected to affect users' decision more. 

In the case of Eco type of persuasion, in particular, the system calculates the total duration the device was in use, from the last time the appliance was turned on till the time the recommendation was created. Based on the given type of each appliance (e.g. lights, A/C unit, etc.) and the list of average CO$_2$ emissions per kWh, as given by the European Environment Agency\footnote{https://www.eea.europa.eu/data-and-maps/indicators/overview-of-the-electricity-production-2/assessment-4 (Last accessed: 01/2020)}, it calculates an estimation of the $CO_2$ emissions for this period of usage.
Likewise, in the case of Econ type of persuasion facts, the system calculates the total cost of the energy consumed by the corresponding appliance over the latest usage period based on the usage time and the average cost of electricity per region, as reported by Eurostat Energy EU\footnote{https://strom-report.de/electricity-prices-europe/ (Last accessed: 01/2020)}. In both cases, we add two additional projections of the consumption in monthly and annual basis. These projection levels are calculated in run-time during the phase of constructing the recommendation and are mainly based on the apriori knowledge of the type of appliance that is being monitor (e.g. LED lights) and the official reports of the average consumption of the type of device and the actual reported electricity costs of the particular area. Combining this knowledge, the total cost of the consumed energy or emissions is calculated given the time of period this device is turned on. For example, let us consider Alice, a typical office user in Greece who turned the A/C as soon as she entered the office on Friday at 09:00 AM. Supposing that the reported consumption of the A/C unit that is installed in Alice's office is 3.2 kW on the cooling mode and given that the electricity price in cents per kWh in Greece is 16.5 cents. If the system decides at 12:00 PM to create a turn-off recommendation for turning the A/C off, then along with the verbal recommendation and the persuasive fact that is chosen it also calculates that for the duration of three hours that the A/C is on the total actual cost so far for this device would be 1.58\EUR. Now based on the knowledge given by the KAM for Alice's consumption habits, if turning on the A/C on Friday mornings is a usual habit the system uses this information and projects the total cost of this habit in a monthly and an annual level, given that Alice would keep on this habit. In the case when the system chooses to include the monthly or annual projection level in the recommendation message, it will help Alice to actually realize the benefit of altering this consumption habit by projecting the total cost of continuing this particular usage of the A/C in a monthly or annual basis.

The use of three different values (the actual value, the monthly and the yearly projection) gives the flexibility to the recommender system, to decide which amount will be presented to the user as a persuasion fact, in order to maximize the probability of acceptance. Providing the monthly or annual amount to the user may in some cases be more informative for the user as it makes it easier for him to comprehend the actual impact of his energy habits either on the environment or his total income.

\section{Results and Discussion}
\label{ss_results}

The experimental process mainly focused on identifying whether the persuasion facts and the explainable recommendations with their more informative content can have a positive impact on the recommendation acceptance by the users. In addition, it is worth identifying if using any level of projection of the total energy savings in the persuasive facts affects the users' decisions on accepting or rejecting a recommendation and thus, affecting their energy footprint.

\begin{figure}[!htp]
 \captionsetup{singlelinecheck = false}
 \centering
  \begin{tikzpicture}[scale=1]
    \begin{axis}[
        width  = 1*\columnwidth,
        height = 5.2cm,
        major x tick style = transparent,
        ybar=1pt,
        bar width=0.8cm,
        ymajorgrids = true,
        ylabel = {Mean acceptance ratio (\%)},
        y label style={at={(axis description cs:-0.08,.5)},anchor=south},
        symbolic x coords={Scenario I,Scenario II,Scenario III},
        xtick = data,
        scaled y ticks = false,
        enlarge x limits=0.5,
        ymin=0,
        ymax= 110,
        legend cell align=left,
        legend style={draw=none, legend columns=-1},
        xtick=data,
        nodes near coords={
            \pgfmathprintnumber[precision=0]{\pgfplotspointmeta}
        }
    ]
    \addplot[style={fill=blue,mark=none},postaction={}, error bars/.cd, y dir=both,y explicit] coordinates {
        (Scenario I, 51) +- (13,13) 
        (Scenario II, 55) +- (21,21) 
        (Scenario III, 70) +- (18,18)};
        \legend{Recommendation Acceptance with STDEV}
    \end{axis}
  \end{tikzpicture}
  \caption{The average acceptance ratio (and standard deviation) of the recommendations for the three scenarios.}\label{total_acceptance_ratio_with_stdev}
\end{figure}
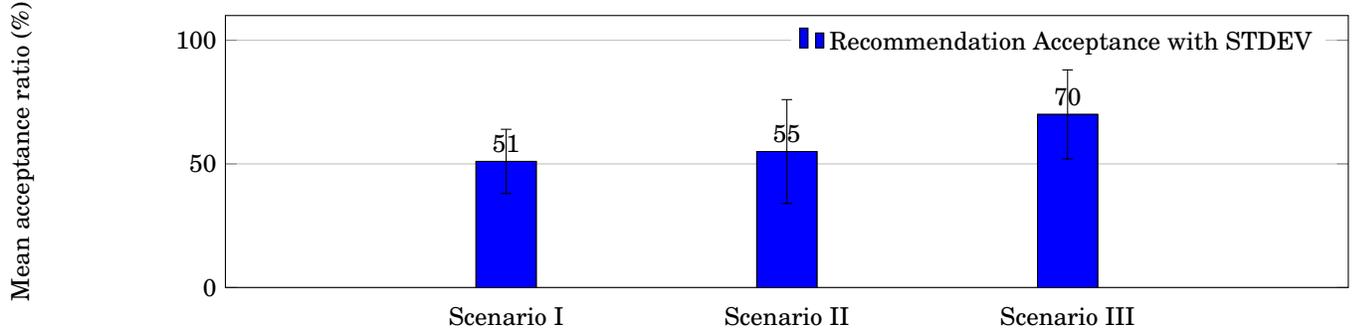

Since a recommendation can be either accepted, rejected or ignored, the definition of recommendation acceptance can either take into account the ignored recommendations or not. Measuring the accepted to rejected recommendations ratio, we have to clarify that the policy definition in the case of ignored recommendations is an important factor. 
In the evaluation scenarios, a follow up recommendation comes after a small interval, whenever the user ignores a recommendation. This interval is bigger whenever the user rejects the recommendation. This means that the total number of recommendations sent to each user is also subject to his/her responses. Overall, almost 16.5\% of the issued recommendations has been ignored, and this number could be different if a different re-issue policy has been followed. In order to avoid the bias of ignored recommendations in the computation of recommendation acceptance, in Figure \ref{total_acceptance_ratio_with_stdev}, we display the average acceptance ratio computed only on the accepted and rejected recommendations for the three scenario versions. Compared to the plain recommendations scenario (I), in Scenario II the persuasive facts increase the average acceptance ratio from 51\% to 55\%. However, the differences between the users are significant and thus the standard deviation is large, which means that the two performances are comparable. On the other side, the performance in Scenario III is almost 19\% higher than the simple scenario, reaching an acceptance ratio of 70\%. The standard deviation around the mean acceptance ratio is smaller, which means that we have a higher degree of agreement between users.

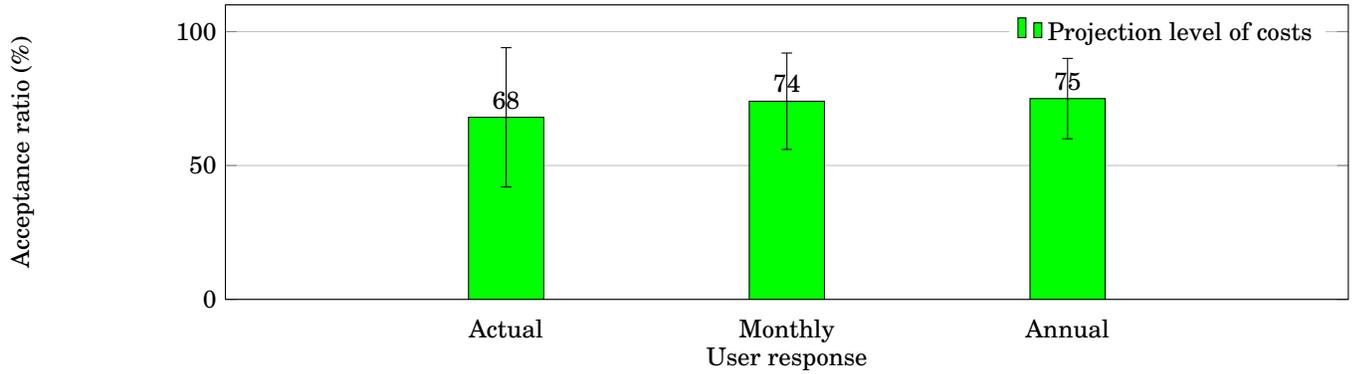
\begin{figure}[!htp]
 \captionsetup{singlelinecheck = false}
 \centering
  \begin{tikzpicture}[scale=1]
    \begin{axis}[
        width  = 1*\columnwidth,
        height = 5.5cm,
        major x tick style = transparent,
        ybar=2pt,
        bar width=1.0cm,
        ymajorgrids = true,
        ylabel = {Acceptance ratio (\%)},
        y label style={at={(axis description cs:-0.08,.5)},anchor=south},
        xlabel = {User response},
        symbolic x coords={Actual,Monthly,Annual},
        xtick = data,
        scaled y ticks = false,
        enlarge x limits=0.5,
        ymin=0,
        ymax= 110,
        legend cell align=left,
        legend style={draw=none, legend columns=-1},
        xtick=data,
        nodes near coords={
            \pgfmathprintnumber[precision=0]{\pgfplotspointmeta}
        }
    ]
        \addplot[style={fill=green,mark=none}, error bars/.cd, y dir=both,y explicit] coordinates {
        (Actual, 68) +- (26,26) 
        (Monthly, 74) +- (18,18) 
        (Annual, 75) +- (15,15) };

        \legend{Projection level of costs}
    \end{axis}
  \end{tikzpicture}
  \caption{Comparison of the total acceptance ratio for the different levels of projection (``Actual", ``Monthly", ``Annual") in the persuasive facts of the recommendations of Scenarios II and III.}\label{acceptance_ratio_aggregation_levels}
\end{figure}

When comparing the average acceptance ratio of the recommendations across the three types of value projections of the persuasion facts (i.e. actual, monthly, yearly), we can see that the report of the actual savings has the worst acceptance ratio (68\%), whereas the acceptance for the monthly and yearly projections is comparable but higher (74\% and 75\% respectively), as depicted in Figure \ref{acceptance_ratio_aggregation_levels}. Also, there is a disagreement between users in the case of persuasive facts that report actual savings, which results in a high standard deviation around the mean in this case.
As an outcome of these results, using either a monthly or an annual projection of the user's consumption costs in the explainable personalized recommendation, makes it easier for the users to be convinced to accept the recommended action created from the system. This makes complete sense, since these projections can be easily understood by the user in terms of realising the total benefit of accepting these recommendations.
These evaluation findings, imply that the combination of the persuasive facts with the explainable illustration of the recommendations, accompanied by a projection of the total cost of the user's energy habits can both assist users easily comprehend the impact of each personalized explainable recommendation and the system to achieve its goal for improving user's energy profile by utilizing well timed personalized turn off actions recommendations.

\begin{figure}[h!]
\centering
\includegraphics[trim=0 0 0 0,clip,width=0.6\columnwidth]{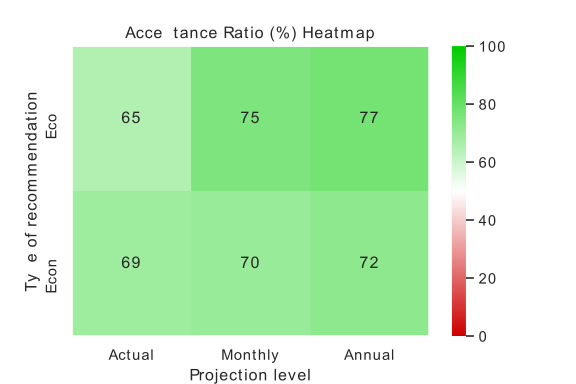}
  \caption{The acceptance ratio heatmap for Scenarios II and III, per type of recommendation (``Eco", ``Econ") and cost projection level (``Actual", ``Monthly", ``Annual").}
\label{fig:acceptance-heatmap}
\end{figure}

Given the increase achieved in the acceptance of the recommendations with the use of explainable recommendations in Scenario III, in Figure \ref{fig:acceptance-heatmap} we compare the acceptance ratio of the recommendations for each combination of projection level used in the recommendations' body (e.g. ``Actual", ``Monthly", ``Annual") and the type of recommendation delivered to the users (e.g. ``Eco" or ``Econ") during the evaluation of the explainable recommender engine (Scenarios II and III) to highlight the optimal combination that mostly managed to trigger users in accepting the recommended action and implicitly achieved the goal of transforming user's energy profile.

As depicted in Figure \ref{fig:acceptance-heatmap}, the use of Eco type of recommendations explanations with an annual projection of total energy cost benefits managed to persuade the users on 77\% of the generated recommendations, whereas the combination of Eco explanations with the use of monthly cost projections also managed to achieve a 75\% of recommendation acceptance. On the other hand, the use of the actual consumption costs at the time the explainable recommendation is triggered combined with an ecological type of persuasion was the least preferred combination for accepting a recommendation with only 65\% of user acceptance ratio.

\section{Conclusions}\label{sec_conclusions}

In this paper, a context-aware explainable recommendation system for energy efficiency is presented. The proposed intelligent system employs a real-time data collection and knowledge extraction and abstraction on real data to generate explainable recommendations, which are personalized to user preferences and habits. 
Explanations are categorized into those that emphasize on the economical saving prospects (Econ) and those that foster a positive ecological impact (Eco). Current results show a 19\% increase on the recommendation acceptance ratio when both economical and ecological persuasive facts are employed. Future work includes enriching the data input to the recommender, employing data visualization as a tool of persuasion, and integration with our mobile application. This work is deemed to be a revolution in recommender system research, where the seeds are planted to develop systems that automatically produce intelligent recommendations for energy saving behavior.

\section{Acknowledgements}\label{acknowledgements}
This paper was made possible by National Priorities Research Program (NPRP) grant No. 10-0130-170288 from the Qatar National Research Fund (a member of Qatar Foundation). The statements made herein are solely the responsibility of the authors.

\end{document}